\documentclass[10pt,twocolumn,letterpaper]{article}

\usepackage[utf8]{inputenc}
\usepackage{booktabs} % 用于生成高质量的表格线 (\toprule, \midrule, \bottomrule)
\usepackage{amsmath}  % 用于支持数学符号，如此处的 \checkmark
\usepackage{booktabs}   % 优美的表格线
\usepackage{pifont}     % 提供 \ding{} 符号（用于 ✓ 和 ✗）
\usepackage[table]{xcolor}  % 支持表格上色（必须），注意加 [table]
\usepackage{colortbl}   % 扩展表格颜色功能（推荐）
\usepackage{graphicx} % 用于调整表格大小（如果需要）

\usepackage[pagenumbers]{cvpr} % To force page numbers, e.g. for an arXiv version

\definecolor{cvprblue}{rgb}{0.21,0.49,0.74}
\usepackage[pagebackref,breaklinks,colorlinks,allcolors=cvprblue]{hyperref}

\def\confName{CVPR}
\def\confYear{2026}

\title{HI-TransPA: Hearing Impairments Translation Personal Assistant}

\author{
Zhiming Ma\thanks{Equal contribution.}\hspace{0.4em}\thanks{Corresponding author. Email: {\tt\small mazhiming312@outlook.com} \& {\tt\small chengshi@shanghaitech.edu.cn}} \\
SmartFlowAI Research\\
Shanghai, China \\
\and
Shiyu Gan\footnotemark[1] \\
Tongji University\\
Shanghai, China \\
\and
Junhao Zhao\footnotemark[1] \\
SmartFlowAI Research\\
Shanghai, China \\
\and
Xianming Li\\
CMIC\\
Guangzhou, China\\
\and
Qingyun Pan\\
BUPT\\
Beijing, China\\
\and
Peidong Wang\\
Northeastern University\\
Shenyang, China \\
\and
Mingjun Pan\\
Peking University\\
Beijing, China \\
\and
Yuhao Mo\\
CMIC\\
Guangzhou, China\\
\and
Jiajie Cheng\\
CMIC\\
Guangzhou, China\\
\and
Chengxin Chen\\
CMIC\\
Guangzhou, China\\
\and
Zhonglun Cao\\
CMIC\\
Guangzhou, China\\
\and
Chonghan Liu\\
Qiyuan Tech \\
Beijing, China \\
\and
Shi Cheng\footnotemark[2]\\
SmartFlowAI Research\\
Shanghai, China \\
}

\begin{document}
\maketitle
\begin{abstract}
Hearing-impaired individuals often face significant barriers in daily communication due to the inherent challenges of producing clear speech. To address this,  we introduce the Omni-Model paradigm into assistive technology and present HI-TransPA, an instruction-driven audio-visual personal assistant. The model fuses indistinct speech with lip dynamics, enabling both translation and dialogue within a single multimodal framework. To address the distinctive pronunciation patterns of hearing-impaired speech and the limited adaptability of existing models, we develop a multimodal preprocessing and curation pipeline that detects facial landmarks, stabilizes the lip region, and quantitatively evaluates sample quality. These quality scores guide a curriculum learning strategy that first trains on clean, high-confidence samples and progressively incorporates harder cases to strengthen model robustness. Architecturally, we employs a novel unified 3D-Resampler to efficiently encode the lip dynamics, which is critical for accurate interpretation. Experiments on purpose-built HI-Dialogue dataset show that HI-TransPA achieves state-of-the-art performance in both literal accuracy and semantic fidelity. Our work establishes a foundation for applying Omni-Models to assistive communication technology, providing an end-to-end modeling framework and essential processing tools for future research.
\end{abstract}    
% ================================================
% Introduction Section (LaTeX version)
% ================================================
\section{Introduction}
\label{sec:intro}

In recent years, the rapid evolution of Large Language Models (LLMs) has driven artificial intelligence into a new era. Building upon this foundation, multimodal models have unified visual, auditory, and textual modalities, demonstrating remarkable capabilities in perception, reasoning, and generation~\cite{llamaomni,qwen2omni,xu2025qwen3,ye2025omnivinci}. These advances have led to significant breakthroughs in areas such as machine translation, human–computer interaction, and content creation. By integrating audio-visual information, machines can now not only ``see'' and ``hear'' but also comprehend complex cross-modal contexts, thereby establishing a robust foundation for tackling real-world communication challenges.

Despite the remarkable progress of artificial intelligence, accessibility and real-world applicability remain limited for certain groups. Individuals with disabilities constitute a substantial share of the global population and still encounter barriers that prevent them from fully benefiting from AI technologies. Encouragingly, early research efforts have begun to bridge this gap by demonstrating how AI can promote greater inclusivity. For instance, vision-based systems such as electronic guide dogs~\cite{InternDog} and OpenAIglasses for Navigation~\cite{AI-FanGe2025OpenAIglasses} have been developed to assist visually impaired users in enhancing their mobility and independence. 
These pioneering works underscore the potential of socially oriented, human-centric AI in promoting accessibility.

Among various disability groups, individuals with hearing loss face particularly intricate and far-reaching challenges. According to the latest estimates from the World Health Organization, more than \textbf{1.5~billion people} worldwide live with some degree of hearing loss, and over \textbf{430~million} require rehabilitative support owing to moderate or higher levels of impairment~\cite{who_wrh_2021}. The impact of hearing loss extends well beyond auditory perception itself. It interferes with natural language acquisition and speech development, which in turn leads to persistent difficulties in verbal communication. These limitations restrict educational and occupational participation and reduce access to information and social interaction. Consequently, hearing loss can result in social isolation, psychological distress, and deepened social inequality.

Most existing assistive technologies focus on converting speech from hearing individuals into text, enabling deaf or hard-of-hearing users to access spoken information. However, these systems offer little support when hearing-impaired users attempt to express themselves. Conventional speech recognition models~\cite{sensevoice,paraformer} are typically trained on standard speech data and therefore struggle to interpret atypical or partially articulated utterances. This limitation hinders the ability of hearing-impaired individuals to engage in spoken interaction and underscores the need for adaptive multimodal models that can understand and process diverse and non-standard speech characteristics.

\begin{figure}[t]
  \centering
  \includegraphics[width=\linewidth]{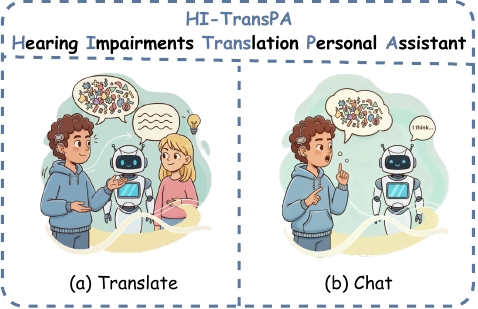}
  \caption{%
  HI-TransPA functions as (a) a translator and (b) an intelligent assistant for individuals with hearing impairments.%
  }
  \label{fig:dataflow}
\end{figure}

To this end, we introduce the Omni-Model paradigm for assistive technologies targeting hearing-impaired users and present HI-TransPA, an instruction-driven audio-visual personal assistant. To the best of our knowledge, this is the first unified framework that integrates \textit{translation} and \textit{chat} functionalities within a single model explicitly designed for assisting the hearing-impaired.
To enable HI-TransPA to learn effectively from such inherently challenging data, we first design a comprehensive data processing pipeline that purifies input signals and then evaluates and partitions the data based on quality. Building upon this partitioning, we then propose a novel quality-aware curriculum learning strategy that leverages the structured data for robust training. Through these methods, HI-TransPA simultaneously processes both raw audio and high-frame-rate visual cues from lip dynamics, and acts as an intelligent conversational partner that understands context and infers user intent. 

Our main contributions are summarized as follows:
\begin{enumerate}
    \item \textbf{Unified audio-visual framework.} We introduce \textbf{HI-TransPA}, an instruction-driven multimodal framework for assisting hearing-impaired individuals. By jointly modeling auditory signals and high-frame-rate lip dynamics, HI-TransPA performs accurate translation and natural dialogue within a single end-to-end architecture.

    \item \textbf{Comprehensive multimodal data pipeline.} We construct a full pipeline covering preprocessing and quality-aware curation. It first extracts and aligns the lip region through a two-stage method, then applies rejection sampling based on multimodal metrics to automatically split data into ``easy'' and ``hard'' subsets.

    \item \textbf{Quality-aware curriculum learning.} We propose a data-driven curriculum strategy that ranks multimodal samples by quality and schedules training from easy to hard, leading to stronger generalization across diverse hearing-impaired speech patterns.
\end{enumerate}
\begin{figure*}[h]
  \centering
  \includegraphics[width=1\linewidth]{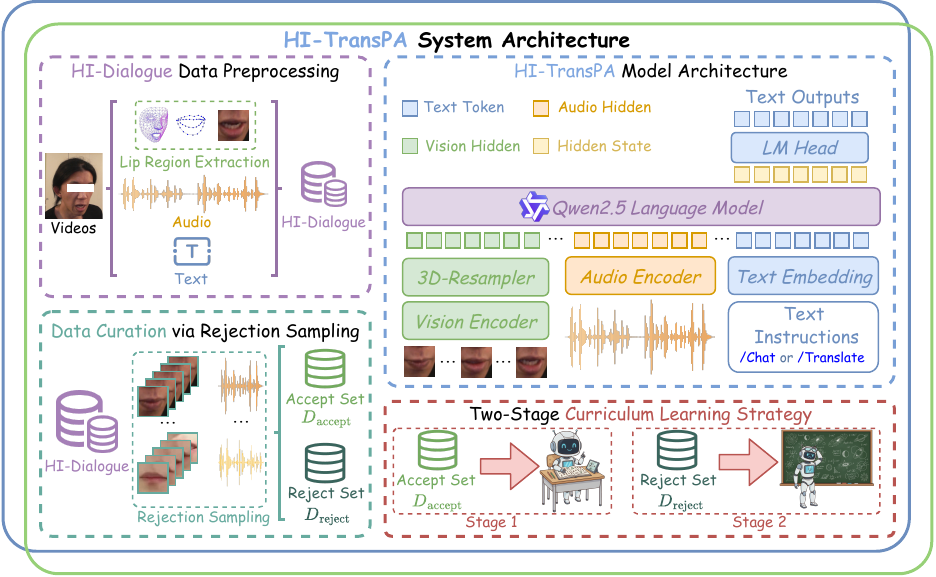}
  \caption{HI-TransPA System Architecture.}
  \label{fig:main}
\end{figure*}

\section{Related Work}

\subsection{ASR and Multimodal Large Language Models}

The field of audio-centric AI has rapidly evolved from narrow, task-specific systems to comprehensive multimodal understanding. Early progress in Automatic Speech Recognition (ASR) was driven by large-scale weakly supervised training (e.g., Whisper~\cite{whisper}), data-efficient modeling (e.g., Canary~\cite{canary}), and architectures optimized for fast inference (e.g., Paraformer~\cite{paraformer}). A major paradigm shift subsequently occurred with the integration of Large Language Models (LLMs), as demonstrated by FireRedASR-AED~\cite{fireredasr}, which enhances ASR performance through greater contextual comprehension and reasoning.

This evolution has given rise to Large Audio-Language Models (LALMs), which treat audio as a rich semantic modality beyond transcription. Representative models such as Qwen2-Audio~\cite{qwen2audio}, MiDashengLM~\cite{midashenglm}, and Step-Audio2\cite{step} exhibit open-domain auditory understanding, supporting instruction following and spoken dialogue across diverse contexts.

Building upon these advances, Omni-Models have emerged to unify text, audio, and visual modalities within a single foundation model. Both proprietary and open-source systems, including GPT-4o~\cite{gpt}, and Qwen2.5-Omni~\cite{qwen2omni}, demonstrate impressive multimodal reasoning capabilities. However, their applications in assistive technology for individuals with hearing loss remain limited. Progress toward inclusive communication tools is further constrained by the scarcity of multimodal datasets, particularly those that containing indistinct speech aligned with lip dynamics.

\subsection{AI for Hearing-Impaired Assistance}

Research on AI-based hearing support has traditionally focused on narrow subtasks, with a predominant emphasis on sign language processing, while the development of tools facilitating speech expression has received comparatively little attention.

\noindent\textbf{Sign Language Recognition and Translation.}
Early studies in sign language understanding focused on isolated sign recognition through multimodal fusion of skeleton keypoints, RGB frames, and depth data~\cite{jiang2021sign}. For continuous sign translation, Camgoz~\textit{et al.}~\cite{camgoz2020sign} introduced an end-to-end joint recognition and translation framework. More recently, contrastive learning has been used to align textual and visual sign embeddings~\cite{jiang2024signclip}, and incorporating non-manual cues such as lip dynamics has been shown to further enhance translation accuracy~\cite{wu2025signclip}.

\noindent\textbf{Lip Reading and Lip-to-Speech Synthesis.}
Complementary lines of work in visual speech processing have focused on decoding speech information directly from lip motions. Cross-modal knowledge distillation enables lip-reading models to exploit representations from pre-trained speech recognizers~\cite{zhao2020hearing}. In addition, recent lip-to-speech synthesis methods reconstruct intelligible speech directly from silent lip videos by leveraging self-supervised discrete speech units as intermediate representations~\cite{choi2023intelligible}.

Despite these promising advances, most approaches still regard individuals with hearing loss as passive recipients of translation output, focusing primarily on perception rather than expression. Enabling active speech articulation is essential to reducing communication barriers, promoting social inclusion, and enhancing self-perception. The emergence of multimodal \textbf{Omni-Models} offers a promising path forward: unified frameworks capable of fine-grained multimodal understanding and context-driven generation. Motivated by this potential, our proposed \textbf{HI-TransPA} leverages the Omni-Model paradigm to realize an intelligent audio-visual assistant capable of both accurate translation and interactive dialogue, supporting truly bidirectional communication for the hearing-impaired.
\section{Method}

To address the two main challenges of noise and heterogeneity in raw data, and the inadequacy of existing Omni-Models for modeling high-frame-rate lip dynamics, we develop \textbf{HI-TransPA}, an Omni-Model with three key components: 
(1) a data preprocessing and curation pipeline, 
(2) a high-speed vision architecture specialized for lip dynamics, and 
(3) a multi-stage training objective based on curriculum learning. 
The overall framework is illustrated in Fig.~\ref{fig:main}.

\subsection{Data Preprocessing and Curation}

\paragraph{Lip Region Extraction.}  
To mitigate head pose variations, irrelevant facial movements, and background noise, we design a two-stage lip region extraction pipeline.

In the first stage, for each video frame $I_t$, a pre-trained landmark detector $F_{lm}$ extracts $N=468$ facial keypoints with coordinates $(x_i, y_i)$, from which we retain only the $M$ points corresponding to the lip region:
\begin{equation}
    P_t = \{ p_i = (x_i, y_i) \mid i \in S_{\text{lips}} \},
\end{equation}
where $\{p_1, ..., p_N\} = F_{lm}(I_t)$ and $S_{\text{lips}}$ denotes the set of lip-related indices.  
These landmarks form a temporal sequence representing lip motion across frames.

In the second stage, we align and stabilize the lip video $V_{\text{lips}}$.  
For frames with valid landmarks $P_t$, we compute bounding boxes $B_t$ and define a uniform crop size $S$ as:
\begin{equation}
    S = \lceil \gamma \cdot \max(\bar{W}, \bar{H}) \rceil_2,
\end{equation}
where $\gamma = 1.2$ is chosen as expansion factor and $\lceil \cdot \rceil_2$ rounds up to the nearest even number.  
Cropping is centered at each frame’s landmark centroid $C_t$, linearly interpolated when landmarks are missing.  
The resulting stabilized video $V_{\text{lips}}$ normalizes head motion and emphasizes lip dynamics relevant to speech articulation.

\begin{figure}[t]
  \centering
  \includegraphics[width=1\linewidth]{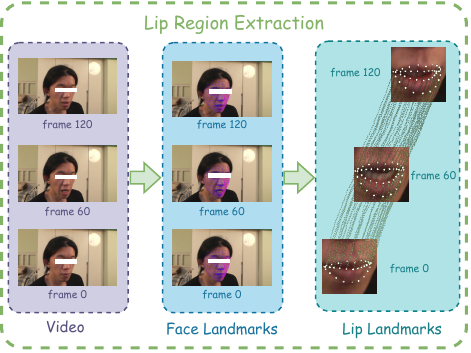}
  \caption{Overview of the lip region extraction pipeline. The left column shows input video frames at different timestamps. In the middle, a face landmark detector identifies 468 facial keypoints (purple). The right column isolates and tracks lip landmarks across frames to produce stabilized lip crops used by the vision encoder.}
  \label{fig:LipRegionExtractionPipeline}
\end{figure}

\begin{figure}[t]
    \centering
    \includegraphics[width=1\linewidth]{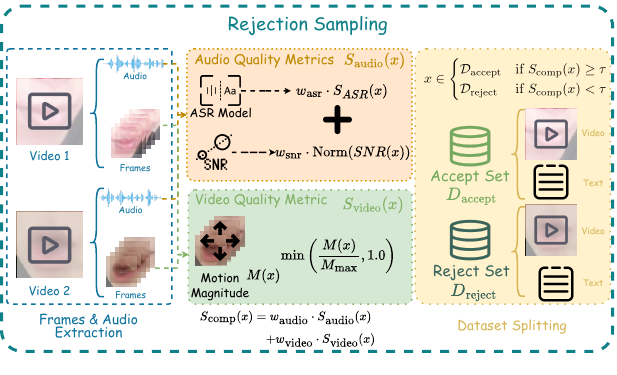}
    \caption{Overview of the rejection sampling pipeline. Each audio-visual sample is scored based on quality metrics from both modalities and then divided into accepted ($\mathcal{D}_{\text{accept}}$) and rejected ($\mathcal{D}_{\text{reject}}$) subsets for curriculum learning.}
    \label{fig:rejection_sampling}
\end{figure}

\paragraph{Data Curation via Rejection Sampling.}  
Even after region extraction, hearing-impaired speech data can exhibit articulation ambiguity or acoustic distortion.  
To ensure data reliability, we introduce a rejection sampling framework that scores each audio-visual pair and partitions the dataset into accepted and rejected subsets, as shown in Fig.~\ref{fig:rejection_sampling}.

\paragraph{Audio Quality.}  
Audio quality is assessed using two complementary metrics.  
The ASR confidence score $S_{\text{ASR}}(x)$ measures the textual consistency between Whisper-large-v3’s transcription $\hat{y}$ and the ground truth $y$:
\begin{equation}
    S_{\text{ASR}}(x) = 1 - \frac{d_{\text{Lev}}(\hat{y}, y)}{\max(|y|, |\hat{y}|)},
\end{equation}
where $d_{\text{Lev}}$ denotes the Levenshtein distance.  
Signal clarity is quantified by the signal-to-noise ratio (SNR), defined as:
\begin{equation}
    \text{SNR}(x) = 10 \cdot \log_{10} \left( \frac{\sum_{t} s_t^2}{\sum_{t} (s_t - \hat{s}_t)^2} \right),
\end{equation}
where $s_t$ represents the clean signal and $\hat{s}_t$ its noisy counterpart.  
A higher SNR indicates clearer speech with less background interference.  
The final composite audio score combines these two aspects:
\begin{equation}
    S_{\text{audio}}(x) = 0.5 \cdot S_{\text{ASR}}(x) + 0.5 \cdot \text{Norm}(SNR(x)),
\end{equation}
where $\text{Norm}(\cdot)$ denotes min-max normalization.

\paragraph{Video Quality.}  
Video quality is characterized by the motion magnitude $M(x)$, defined as the mean pixel difference between consecutive frames.  
We normalize this metric into the range $[0, 1]$:
\begin{equation}
    S_{\text{video}}(x) = \min\left(\frac{M(x)}{M_{\text{max}}}, 1.0\right),
\end{equation}
where $M_{\text{max}}$ is the 90th-percentile motion magnitude across the dataset.

\paragraph{Rejection Sampling.}  \label{sec:rejection_sampling}
The final composite sample score is defined as:
\begin{equation}
    S_{\text{comp}}(x) = 0.6 \cdot S_{\text{audio}}(x) + 0.4 \cdot S_{\text{video}}(x),
\end{equation}
and samples are partitioned as:
\begin{equation}
    x \in
    \begin{cases}
        \mathcal{D}_{\text{accept}}, & \text{if } S_{\text{comp}}(x) \geq 0.55,\\
        \mathcal{D}_{\text{reject}}, & \text{otherwise.}
    \end{cases}
\end{equation}
Rejected samples are later reused as hard examples in the curriculum learning stage to enhance model robustness.

\subsection{Model Architecture}

The proposed \textbf{HI-TransPA} builds upon the Qwen2.5-Omni-3B framework, with a re-engineered vision subsystem specifically optimized for high-frame-rate lip reading.  
The main architectural innovations are the integration of the SigLIP Vision Transformer and the Unified 3D-Resampler module from MiniCPM-V~4.5~\cite{minicpm-v}.

\paragraph{Vision Encoder.} The SigLIP encoder~\cite{siglip} provides fine-grained visual representations suitable for lip articulation modeling.  
It processes the lip video $V_{\text{lips}} \in \mathbb{R}^{T \times H \times W \times C}$ and encodes it into a sequence of patch tokens $Z_{\text{patch}}$.

\paragraph{Spatiotemporal Resampler.} To reduce computation for long video sequences, we apply the Unified 3D-Resampler after the vision encoder.  
It uses cross-attention with $N_q = 64$ learnable queries to compress the token sequence:
\begin{equation}
    Z_{\text{fused}} = \text{3D-Resampler}(Z_{\text{patch}}),
\end{equation}
yielding $Z_{\text{fused}} \in \mathbb{R}^{N_q \times D_{\text{llm}}}$, which preserves essential spatiotemporal cues while reducing token length.

\subsection{Multi-Stage Alignment and Fine-Tuning}

With the vision subsystem redesigned, we perform a three-stage alignment and adaptation process to ensure multimodal synergy between vision, audio, and language.

\paragraph{Stage 1: General Visual Alignment.}  
This stage introduces the language model to the new visual features through two phases:  
(1) \emph{Image alignment}: train only the 3D-Resampler on the Chinese-LLaVA-Vision dataset~\cite{chinese-llava2023} while freezing the vision encoder and LLM;  
(2) \emph{Video alignment}: urther train the Resampler on 30\% of the LLaVA-Video-178K dataset~\cite{zhang2024videoinstructiontuningsynthetic} to capture temporal dynamics.

\paragraph{Stage 2: Audio-Visual Co-adaptation.}  
Using the Chinese-LiPS dataset~\cite{zhao2025chineselipschineseaudiovisualspeech}, we jointly fine-tune the 3D-Resampler and the audio encoder so that both modalities produce complementary embeddings optimized for Audio-Visual Speech Recognition (AVSR).

\paragraph{Stage 3: Conversational Fine-Tuning.}  

We perform end-to-end instruction fine-tuning to jointly optimize the model's translation and dialogue abilities. A mixed-instruction dataset is constructed with two modes: \texttt{/translate}, using paired audio-visual samples and their reference texts; and \texttt{/chat}, which shares the same inputs but pairs them with language-model–generated conversational replies.  
During training, both modes are combined under the curriculum learning strategy described in Section~\ref{sec:training}, where sample difficulty is determined by translation quality. The unified fine-tuning stage enables HI-TransPA to consolidate comprehension and transcription skills while adapting to real conversational scenarios.

\subsection{Training Objective}
\label{sec:training}

To enhance robustness and training stability when learning from multimodal data, we apply a two-stage curriculum learning approach.

\paragraph{Stage 1: Foundational Learning on Accepted Data.}  
We start by fine-tuning the model on high-quality samples $\mathcal{D}_{\text{accept}}$ for three epochs using the cross-entropy loss:
\begin{equation}
    \mathcal{L}_{\text{Stage 1}} = \mathbb{E}_{x \in \mathcal{D}_{\text{accept}}} [L_{\text{CE}}(f(x), y)],
\end{equation}
which aim to establish stable multimodal alignment between lip motion, audio, and textual content.

\paragraph{Stage 2: Robustness Enhancement on Rejected Data.}  
We then continue training for five epochs on the harder subset $\mathcal{D}_{\text{reject}}$:
\begin{equation}
    \mathcal{L}_{\text{Stage 2}} = \mathbb{E}_{x' \in \mathcal{D}_{\text{reject}}} [L_{\text{CE}}(f(x'), y')],
\end{equation}
which is designed to implicitly up-weights challenging examples, encouraging the model to learn robust representations that generalize to noisy, real-world conditions.

Overall, this easy-to-hard learning schedule stabilizes convergence and yields a model capable of consistent performance under diverse acoustic and visual quality levels.

\subsection{Metrics Definition}

\paragraph{Character Error Rate (CER).}  
For evaluation, we report the CER to measure literal transcription accuracy.  
Given a ground-truth sequence $y$ and model output $\hat{y}$, CER is defined as:
\begin{equation}\label{eq:cer}
    \text{CER} = \frac{d_{\text{Lev}}(\hat{y}, y)}{|y|},
\end{equation}
where $d_{\text{Lev}}$ is the Levenshtein distance counting substitutions, deletions, and insertions.  
A lower CER indicates higher transcription fidelity, while a higher CER reflects poorer recognition performance.

\paragraph{Embedding Similarity (EmbSim).}  
To measure the alignment consistency between predicted and reference utterances in embedding space, we compute Embedding Similarity using cosine similarity.  
Given audio-visual feature embeddings $\mathbf{e}_{\text{pred}}$ and $\mathbf{e}_{\text{ref}}$, EmbSim is defined as:
\begin{equation}
\label{eq:embsim}
    \text{EmbSim} = \frac{\mathbf{e}_{\text{pred}} \cdot \mathbf{e}_{\text{ref}}}{\|\mathbf{e}_{\text{pred}}\|_2 \, \|\mathbf{e}_{\text{ref}}\|_2}.
\end{equation}
Higher EmbSim values indicate closer semantic alignment between the predicted representation and the reference embedding, reflecting stronger cross-modal coherence.
\section{Experiments}

% 定义符号
\newcommand{\cmark}{\ding{51}}  % ✓
\newcommand{\xmark}{\ding{55}}  % ✗

% 定义浅灰背景色
\definecolor{lightgrayrow}{gray}{0.95}

\begin{table*}[t]
\centering
\caption{Comparison of various models on the HI-Dialogue test set.}
\label{tab:ai_models}
\begin{tabular}{lcccccc}
\toprule[1.5pt]
\textbf{Models} & \textbf{Params} & \textbf{Audio} & \textbf{Video} & \textbf{EmbSim} & \textbf{CER} & \textbf{CS} \\
\midrule

% -------------------- ASR Models --------------------
\multicolumn{7}{c}{\textit{ASR Models}} \\
\midrule
Whisper-large-V3              & 1.6B & \cmark & \xmark & 0.79 & 0.32 & 0.74 \\

SenseVoice-small              & 0.3B & \cmark & \xmark & 0.71 & 0.35 & 0.68 \\

Paraformer-large              & 0.2B & \cmark & \xmark & 0.70 & 0.38 & 0.66 \\

FireRedASR-AED                & 1.1B & \cmark & \xmark & 0.77 & 0.38 & 0.70 \\

% -------------------- LALMs --------------------
\midrule
\rowcolor{white}
\multicolumn{7}{c}{\textit{LALMs}} \\
\midrule
Qwen2-Audio                   & 7B   & \cmark & \xmark & 0.74 & 0.44 & 0.65 \\
MiDashengLM                   & 7B   & \cmark & \xmark & 0.67 & 0.53 & 0.57 \\
InternLM-XComposer2.5-OmniLive& 7B   & \cmark & \xmark & 0.67 & 0.54 & 0.57 \\
Step-Audio 2 mini             & 7B   & \cmark & \xmark & 0.79 & 0.34 & 0.73 \\

% -------------------- Omni-Models --------------------
\midrule
% \rowcolor{lightgrayrow}
\multicolumn{7}{c}{\textit{Omni-Models}} \\
\midrule
Qwen2.5-Omni (3B)                    & 3B   & \cmark & \cmark & 0.73 & 0.44 & 0.65 \\
Qwen2.5-Omni (7B)                    & 7B   & \cmark & \cmark & 0.75 & 0.42 & 0.67 \\
\textbf{HI-TransPA}                  & 3B   & \cmark & \cmark & 0.77 & 0.37 & 0.70 \\
\textbf{HI-TransPA (Curriculum Learning)} & 3B & \cmark & \cmark & \textbf{0.84} & \textbf{0.27} & \textbf{0.79} \\

\bottomrule[1.5pt]
\end{tabular}
\end{table*}

\subsection{Experimental Setup}

\noindent\textbf{Dataset.} 
We collected and curated a dedicated dataset, HI-Dialogue, to support the training and evaluation of audio-visual dialogue models for hearing-impaired individuals. Six volunteers with varying levels of hearing loss recorded audio-visual materials covering daily conversations, instructional texts, and emergency situations. After manual screening to remove samples with occluded lip regions or mismatched transcripts, we obtained 9,673 high-quality samples.

The dataset is split into 7,736 training and 1,937 test samples (80/20). Following the rejection sampling strategy (Sec.~\ref{sec:rejection_sampling}), the training data is further divided into an \textbf{accepted set} ($\mathcal{D}_{\text{accept}}$, 4,733 samples) and a \textbf{rejected set} ($\mathcal{D}_{\text{reject}}$, 3,003 samples) used for curriculum learning. To enhance conversational capability, text responses are distilled for each training sample to support instruction tuning.

\subsection{Evaluation Metrics}
\label{sec:metrics}

We evaluate the models primarily on their comprehension accuracy.  
To quantify both literal and semantic correctness, we define a \textbf{Comprehensive Score (CS)} as:
\begin{equation}
    \text{CS} = (1 - \alpha) \cdot (1 - \text{CER}) + \alpha \cdot \text{EmbSim},
    \label{eq:cs}
\end{equation}
and $\alpha = 0.5$ is adopted in our experiments.  

The \textbf{Character Error Rate (CER)} follows the definition in Eq.~\ref{eq:cer} of Sec.~\ref{sec:training}; it measures literal transcription accuracy, where a lower value indicates better performance and a higher value denotes more transcription errors.  
The \textbf{Embedding Similarity (EmbSim)}, defined in Eq.~\ref{eq:embsim}, is computed using the cosine similarity between the predicted and reference embeddings, where higher values indicate stronger semantic consistency.  
We adopt Qwen3-Embedding-0.6B~\cite{qwen3embedding} for embedding extraction.

This balanced formulation in Eq.~\ref{eq:cs} provides a holistic measure that reflects both surface-level accuracy (through CER) and deeper semantic fidelity (through EmbSim).  
We omit conversational fluency and empathy metrics to maintain comparability among models with different dialogue generation capabilities, focusing instead on the upstream comprehension of noisy audio-visual signals—a shared foundation across all systems evaluated.

\subsection{Baselines}
\label{sec:baselines}

We benchmark \textbf{HI-TransPA} against a set of representative speech and multimodal systems, covering three major categories as summarized in Table~\ref{tab:ai_models}. All baseline models are fine-tuned on the HI-Dialogue dataset for comparability. Together, they form a comprehensive spectrum ranging from traditional audio-only recognizers to state-of-the-art multimodal Omni-Models.

\noindent\textbf{ASR Models.}  
We select four widely used automatic speech recognition (ASR) systems representing strong audio-only foundations with different design philosophies. 
Whisper-large-v3~\cite{whisper} is trained on 680k hours of weakly supervised web data and serves as a robust multilingual reference under zero-shot conditions.
SenseVoice-small~\cite{sensevoice} is part of the FunAudioLLM framework, supporting fast multilingual ASR and audio-event recognition with low latency.
Paraformer-large~\cite{paraformer} employs a non-autoregressive parallel Transformer to accelerate inference while maintaining competitive accuracy.
FireRedASR-AED~\cite{fireredasr} represents an industrial-grade, Mandarin-optimized encoder–decoder architecture integrating large language modeling for superior character error rate (CER) performance.
These models quantify the upper bound of purely acoustic transcription capability and provide a foundation for measuring the gains from incorporating visual inputs.

\noindent\textbf{Large Audio Language Models (LALMs).}  
To evaluate semantic reasoning on spoken content, we benchmark against large audio–language models that couple LLM backbones with acoustic encoders, yet remain unimodal. 
Qwen2-Audio~\cite{qwen2audio} enables direct comprehension of complex auditory scenes through instruction-following and conversational understanding.
MiDashengLM~\cite{midashenglm}, InternLM-XComposer2.5-OmniLive~\cite{internlm}, and Step-Audio~2~mini represent different scales of audio-centric systems, combining pretrained LLMs with efficient auditory front-ends for contextual reasoning. 
While these models excel in semantic comprehension, they cannot exploit visual cues—such as lip dynamics—to disambiguate indistinct or impaired speech, which is the focus of our proposed system.

\noindent\textbf{Omni-Models.}  
Finally, we compare with open-source multimodal Omni-Models that jointly process audio, visual, and textual information.
Qwen2.5-Omni (3B and 7B)~\cite{qwen2omni} integrates synchronized audio–visual encoders and a shared language backbone to handle streaming multimodal tasks, representing the state of general-purpose cross-modal LLMs.
Our HI-TransPA and its variant trained with curriculum learning follow the Omni-Model paradigm but are specifically optimized for accessibility. 
By aligning high-frame-rate lip features and audio inputs through instruction tuning, HI-TransPA demonstrates improvements in both embedding similarity and character error rate, outperforming generic Omni baselines by a clear margin.

\begin{figure}[t]
    \centering
    \includegraphics[width=1\linewidth]{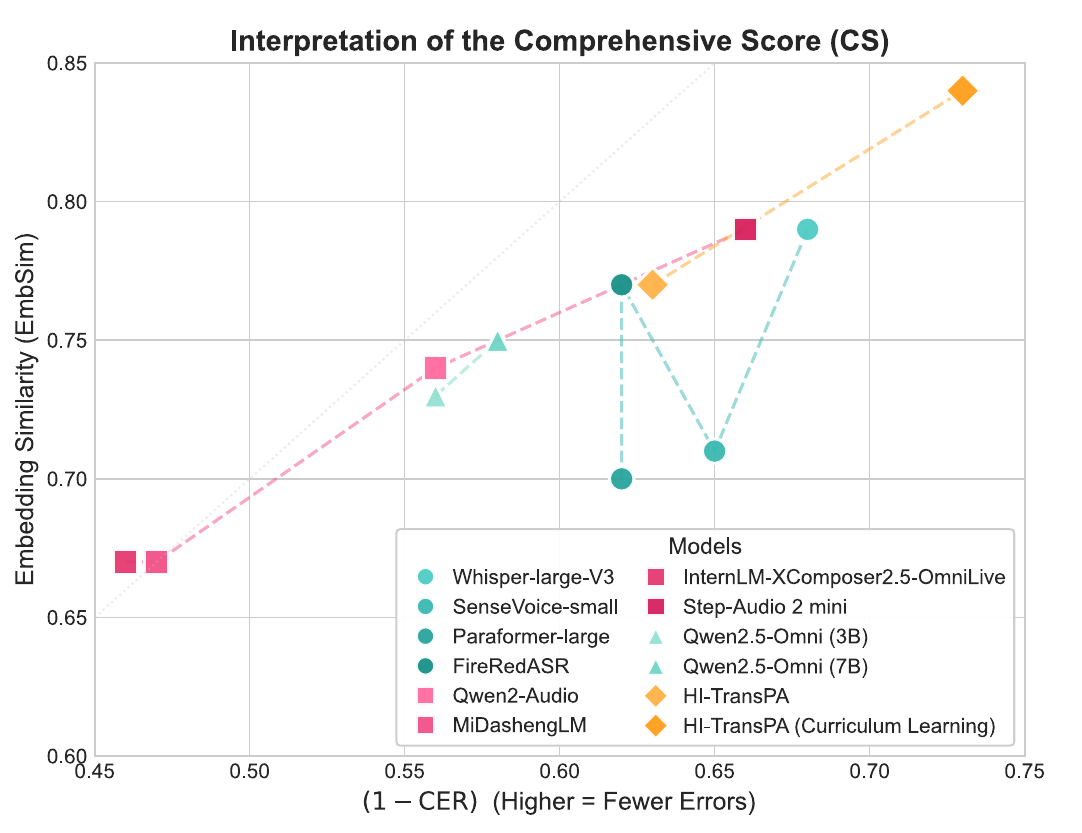}
    \caption{Interpretation of the Comprehensive Score (CS).  
    Models are plotted in the $(1-\text{CER})$ vs. EmbSim space.  
    Higher positions and rightward movement correspond to lower error rates and stronger semantic consistency, jointly yielding higher CS values.}
    \label{fig:cs_interpret}
\end{figure}

\subsection{Main Results and Analysis}

The detailed comparison results are presented in Table~\ref{tab:ai_models}.

\noindent\textbf{Baseline analysis.}  
Audio-only models perform significantly worse on hearing-impaired speech: Whisper-large-V3 and Step-Audio 2 mini achieve CS scores of 0.74 and 0.73 but maintain high CERs (Eq.~\ref{eq:cer}).  
Adding general vision encoders, as in Qwen2.5-Omni, yields limited gains (CS = 0.67), suggesting that generic multimodal fusion fails to capture the fine-grained lip dynamics.

\noindent\textbf{Effectiveness of HI-TransPA.}  
Our 3B HI-TransPA already surpasses the larger 7B Qwen2.5-Omni, achieving a CS of 0.70.  
The specialized vision architecture, optimized for high-frame-rate lip motion, proves crucial.  
Incorporating curriculum learning further boosts performance, yielding a CS of 0.79, an EmbSim of 0.84 (Eq.~\ref{eq:embsim}), and a reduced CER of 27\%—the best among all baselines.

\subsection{Interpretation of the Comprehensive Score}

The comprehensive score (Eq.~\ref{eq:cs}) combines literal correctness $(1-\text{CER})$ and semantic coherence (EmbSim), enabling us to compare models beyond recognition accuracy. From Table~\ref{tab:ai_models}, three core observations emerge:

\begin{enumerate}[leftmargin=1.2em]
    \item \textbf{ASR models:}  
    Whisper-large-V3 shows a moderate CER (\(32\%\)) and EmbSim = 0.79, resulting in CS = 0.74.  
    This indicates that despite high transcription accuracy, audio-only models lack multimodal semantic grounding.
    
    \item \textbf{Generic Omni-models:}  
    Qwen2.5-Omni (3B and 7B) achieve modest improvement.  
    They benefit from visual cues but fail to capture fine-grained temporal lip dynamics, keeping both CER and EmbSim suboptimal.
    
    \item \textbf{HI-TransPA:}  
    Our model enhances both dimensions simultaneously, reducing the CER from 42\% to 27\% and increasing the EmbSim from 0.75 to 0.84, which leads to a CS of 0.79.  
    This indicates that HI-TransPA not only produces more accurate transcripts but also better preserves contextual meaning through specialized lip-motion modeling and curriculum-based training.
    
\end{enumerate}

Fig.~\ref{fig:cs_interpret} visualizes different models on a two-dimensional plane defined by $(1-\text{CER})$ and EmbSim. 
Points closer to the upper-right corner indicate a better balance between literal accuracy and semantic coherence. 
The distribution shows that the proposed Comprehensive Score (CS) effectively reflects each model’s overall comprehension ability by jointly capturing text-level correctness and meaning preservation, allowing direct comparison across architectures with different modalities and training paradigms.

\begin{figure}[t!]
    \centering
    \includegraphics[width=1\columnwidth]{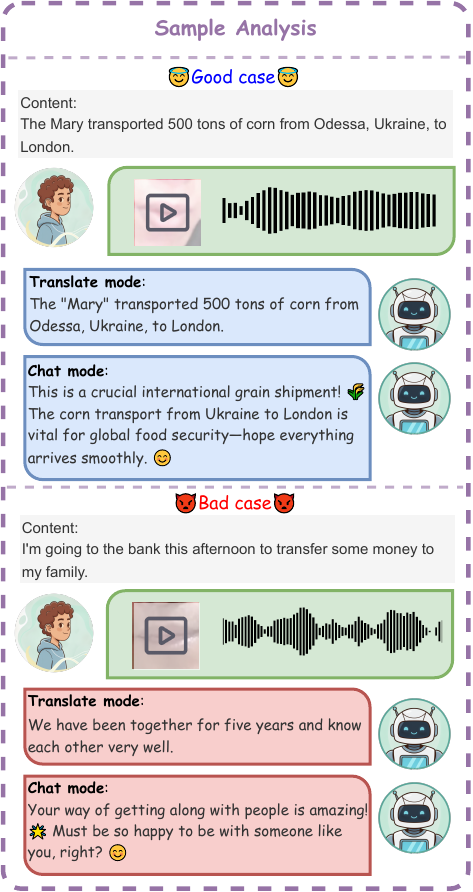} 
    \caption{Qualitative examples illustrating the importance of comprehension. The ``Good case'' (from our HI-TransPA) shows how accurate translation enables a relevant and helpful chat response. The ``Bad case'' (from a baseline) demonstrates how a failure in comprehension leads to a completely irrelevant dialogue, underscoring our ``Comprehension First'' evaluation principle.}
    \label{fig:sample_analysis}
\end{figure}

\subsection{Ablation Studies}

\noindent\textbf{Role of the Visual Modality.}  
To assess the contribution of visual input, we compare the full model with an audio-only variant.  
As shown in Table~\ref{tab:ablation_vision}, removing the visual modality severely degrades performance (CS drops from 0.70 to 0.64; CER rises from 37\% to 46\%, see Eq.~\ref{eq:cer}), confirming that lip motion provides indispensable cues for comprehension.

\noindent\textbf{Effect of Curriculum Learning.}  
Finally, we analyze the impact of the proposed two-stage training strategy (Sec.~\ref{sec:training}).  
Compared with the base model trained without curriculum learning, the final version improves CS (Eq.~\ref{eq:cs}) from 0.70 to 0.79 and reduces CER (Eq.~\ref{eq:cer}) from 37\% to 27\%, confirming that progressive training enhances model robustness against noisy and difficult samples.

\subsection{Qualitative Analysis}
\label{sec:qualitative_analysis}

To provide a more intuitive demonstration of our model's capabilities and to analyze the critical role of comprehension accuracy in generating meaningful dialogue, we present two representative cases in Figure~\ref{fig:sample_analysis}. These cases showcase two modes of operation for HI-TransPA: a \textbf{``Translate mode,''} which aims for a literal transcription that reveals the model's true understanding of the input, and a \textbf{``Chat mode,''} which aims to generate a natural and helpful conversational response based on that understanding.

\begin{table}[ht]
\centering
\caption{Ablation on the effectiveness of the visual modality.}
\label{tab:ablation_vision}
\begin{tabular}{lccccc}
\toprule[1.5pt]
Model & Audio & Video & EmbSim & CER(\%) & CS \\
\midrule
Our(A) & \checkmark & & 0.74 & 46 & 0.64 \\
Our(O) & \checkmark & \checkmark & 0.77 & 37 & 0.70 \\
\bottomrule[1.5pt]
\end{tabular}
\end{table}

In the ``Good case,'' our HI-TransPA model successfully processes a complex sentence containing multiple entities. The output in ``Translate mode'' is a near-perfect transcription, demonstrating its robust multimodal comprehension. Based on this precise understanding, the ``Chat mode'' is able to go beyond simple repetition, infer the broader context of an ``international grain shipment,'' and provide an empathetic and insightful response, truly functioning as an intelligent assistant.

In contrast, the ``Bad case'' vividly illustrates what occurs when comprehension fails (this result is from a poorly performing baseline). The ``Translate mode'' output reveals a complete misinterpretation of the user's intent, producing a sentence semantically unrelated to the original input about transferring money at a bank. This fundamental failure in comprehension directly explains why the ``Chat mode'' response is nonsensical in context. Although the chat response is grammatically coherent, it is logically derived from an incorrect premise, rendering it entirely useless and potentially confusing for the user.

The comparison strongly validates our \textbf{Comprehension First} principle (Section~\ref{sec:metrics}), showing that accurate comprehension, reflected in literal translation, forms the basis of all higher-level dialogue abilities. As confirmed in Table~\ref{tab:ai_models}, HI-TransPA’s superior understanding establishes a solid foundation for meaningful and supportive interaction.

\section{Conclusion}

We presented \textbf{HI-TransPA}, an instruction-driven audio-visual Omni-Model that designed to assists people with hearing loss in natural dialogue and precise speech translation. With jointly modeling speech and high-frame-rate lip cues with a redesigned visual encoder and a quality-aware curriculum strategy for training, HI-TransPA achieves outstanding performance on multimodal real-world scenarios. Extensive experimental results on the HI-Dialogue dataset demonstrate that HI-TransPA consistently outperforms existing ASR methods, LALMs, and general Omni-Models, highlighting the potential of instruction-tuned multimodal intelligence for inclusive communication.

\newpage
{
    \small
    \bibliographystyle{ieeenat_fullname}
    \bibliography{main}
}

\end{document}